\pdfoutput=1

\documentclass[11pt]{article}

\usepackage[final]{acl}

\usepackage{times}
\usepackage{latexsym}

\usepackage[T1]{fontenc}

\usepackage[utf8]{inputenc}

\usepackage{microtype}

\usepackage{inconsolata}

\usepackage[T1]{fontenc}    
\usepackage{hyperref}       
\usepackage{url}            
\usepackage{booktabs}       
\usepackage{amsfonts}       
\usepackage{nicefrac}       
\usepackage{microtype}      
\usepackage{xcolor}         
\usepackage{amsmath}
\usepackage{times}
\usepackage{latexsym} 
\usepackage{amssymb}
\usepackage{graphicx}
\usepackage{subcaption}
\usepackage{booktabs}
\usepackage{listings}
\usepackage{xcolor}
\usepackage{wrapfig}
\usepackage{pgfplots}
\usepackage{pgfplotstable}
\pgfplotsset{compat=1.18}

\definecolor{codeblue}{rgb}{0.25,0.5,0.75}
\definecolor{codegreen}{rgb}{0.0,0.6,0.0}
\definecolor{codegray}{rgb}{0.5,0.5,0.5}
\definecolor{codepurple}{rgb}{0.58,0.0,0.82}

\lstdefinestyle{mystyle}{
    language=Python,
    basicstyle=\ttfamily\small,
    keywordstyle=\color{codeblue}\bfseries,
    commentstyle=\color{codegreen}\itshape,
    stringstyle=\color{codepurple},
    numberstyle=\tiny\color{codegray},
    breaklines=true,
    mathescape=true,
    showstringspaces=false,
    frame=none,
    tabsize=4
}
\title{Differential Mamba}

\author{
Nadav Schneider$^{1,2}$,
Itamar Zimerman$^{3,4}$,
Eliya Nachmani$^{1}$ \\
\\
$^{1}$School of Electrical and Computer Engineering, Ben-Gurion University of the Negev \\
$^{2}$IAEC \\
$^{3}$Tel-Aviv University \\
$^{4}$IBM Research \\
\texttt{nadavsch@post.bgu.ac.il} \quad
\texttt{zimerman1@mail.tau.ac.il} \quad
\texttt{eliyanac@bgu.ac.il}
}

\begin{document}
\maketitle
\begin{abstract}
Sequence models like Transformers and RNNs often overallocate attention to irrelevant context, leading to noisy intermediate representations. This degrades LLM capabilities by promoting hallucinations, weakening long-range and retrieval abilities, and reducing robustness. Recent work has shown that differential design can mitigate this issue in Transformers, improving their effectiveness across various applications. In this paper, we explore whether these techniques, originally developed for Transformers, can be applied to Mamba, a recent architecture based on selective state-space layers that achieves Transformer-level performance with greater efficiency. We show that a naive adaptation of differential design to Mamba is insufficient and requires careful architectural modifications. To address this, we introduce a novel differential mechanism for Mamba, empirically validated on language modeling benchmarks, demonstrating improved retrieval, long-context capabilities
and superior performance over vanilla Mamba. Finally, we conduct extensive ablation studies and empirical analyses to justify our design choices and provide evidence that our approach effectively mitigates the overallocation problem in Mamba-based models. Our code is publicly available.

\vspace{0.5em}
\hspace{.5em}
\includegraphics[width=1.25em,height=1.15em]{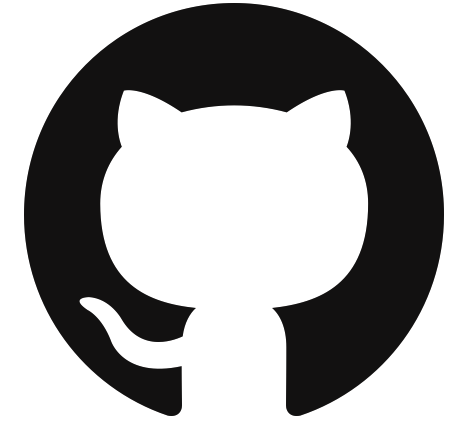}\hspace{.75em}
\parbox{\dimexpr\linewidth-7\fboxsep-7\fboxrule}{\url{https://github.com/NadavSc/Diff-Mamba}}
\vspace{-.5em}
\end{abstract}

\section{Introduction\label{sec:intro}}
Designing enhanced architectures for deep sequence modeling is a pivotal task in the ML community, as sequence models drive breakthroughs such as ChatGPT~\citep{achiam2023gpt} in NLP and Stable diffusion~\citep{rombach2022high} in computer vision, which are fundamental to modern generative models. However, these models face two major challenges as: efficiency, constrained by the quadratic time complexity in sequence length of the Transformer architecture~\citep{vaswani2017attention}, and robustness, which is hindered by inconsistency, reliability issues, and hallucinations, leading to sub-optimal performance. Our work addresses these challenges by enhancing the robustness of recent efficient architectures with sub-quadratic complexity such as Mamba~\citep{gu2024mamba}, making them more reliable and robust.

To address this robustness issue, we aim to reduce the \textbf{over-allocation of attention to irrelevant context} across hidden layers, which often leads to noisy representations. Our approach is inspired by~\citet{ye2024differential}, who mitigated this problem in transformers through differential design. This architectural modification was the core of the Diff-Transformer, a transformer variant that demonstrated improved 
performance, including greater robustness, enhanced retrieval and long-context capabilities compared to the original model.

Our focus is on improving the recently introduced Mamba architecture, which builds upon selective state-space layers (S6). This architecture is known for its efficiency, introducing sub-quadratic complexity in sequence length while also enabling auto-regressive decoding with a complexity that remains independent of sequence length. Beyond these efficiency advantages, recent studies have demonstrated that Mamba-based architectures can match or even surpass the SoTA performance of transformers, even at relatively large scales. For example, Falcon-Mamba 7B~\citep{zuo2024falcon}, a pure Mamba model, matches the performance of LLaMA- 3-8B on several language tasks. Additional notable examples of Mamba’s integration in LLMs include Jamba~\citep{lieber2024jamba}, Zamba~\citep{glorioso2024zamba}, Hymba~\citep{dong2024hymba}, and others~\citep{wang2024mambabyte,waleffe2024empirical,ren2024samba}. %
More importantly, recent Mamba-based models have demonstrated remarkable performance as reasoning models~\citep{paliotta2025thinking,wang2025m1}, underscoring their central role in the ongoing test-time scaling revolution, as exemplified by models such as OpenAI’s O1~\citep{jaech2024openai} and DeepSeek’s R1~\citep{guo2025deepseek}.


While the over-allocation problem is a general issue not specific to any architecture, we hypothesize that Mamba-based LLMs have a stronger tendency toward over-allocation compared to transformers. This is primarily due to two factors: (i) Mamba is a softmax-free architecture, meaning it lacks the exponential scaling effect that helps suppress irrelevant attention weights, and (ii) as a state-based model, Mamba operates locally and cannot directly process distant tokens without considering all intermediate tokens, resulting in the dispersion of important tokens among irrelevant ones. This leads to our central research question:

\smallskip
{\centering\textit{Can differential design be leveraged to improve the robustness of Mamba models?}\par}
\noindent where we hypothesize that improving robustness by mitigating the over-allocation problem can enhance the model’s general capabilities, improve retrieval and long-context processing, and potentially reduce issues related to consistency and hallucinations.

We provide a positive answer to this question by showing that, although a naive implementation of differential mechanisms does not improve Mamba architectures, a more carefully designed mechanism does. Through a systematic evaluation on language tasks, ablation studies, and empirical analysis, we conclude that our variant is favorable compared to the vanilla Mamba.

\smallskip
\textbf{Our main contributions} are as follows: (i) We present Diff-Mamba, a modification of the Mamba architecture inspired by Diff-Transformer, which mitigates the problem of over-allocating attention scores to irrelevant context and improves the general language modeling abilities of the model.
(ii) Through a series of ablation studies and empirical analysis using mechanistic interpretability tools, we justify our design choices and demonstrate that the intermediate representations obtained from our method are indeed less noisy. (iii) Finally, we show that Diff-Mamba demonstrates improved retrieval and long-context capabilities compared to Mamba. This is particularly important, as recurrent LLMs such as Mamba are primarily designed to address the inefficiencies caused by the quadratic complexity of transformers, which becomes especially critical in long-context regimes. Accordingly, Diff-Mamba achieves improvements over Mamba precisely in the domain where such architectures are most needed~\citep{ben2025overflow}.

\section{Background\label{sec:background}}
Here we describe the scientific context and introduce the terminology for discussing our method.

\subsection{Differential Transformer}
\paragraph{Self-Attention}
Self-Attention is a fundamental component of Transformer architectures~\citep{vaswani2017attention}, has significantly shaped recent advances in both NLP and computer vision. This mechanism enables dynamic focus allocation by capturing pairwise token dependencies, allowing the model to determine the relative importance of each token within a sequence. Mathematically, it defined by:

\vspace{-15pt}
{\small
\begin{equation}
\label{eq:attnMAT}
\text{Attn}(Q, K, V) = \alpha V, \quad \alpha = \text{softmax}\left(\frac{QK^T}{\sqrt{d_k}}\right) 
\end{equation}
}
In this formulation, $Q$,$K$, and $V$ represent the queries, keys, and values, respectively, while $d_k$ denotes the key dimension. Transformers extend this mechanism by employing $H$ parallel attention heads, enabling the model to capture a broader spectrum of dependencies.

\paragraph{Differential Attention}
To address the problem of over-allocation of attention to irrelevant tokens, \citet{ye2024differential}, introduced Diff-Transformer, a mechanism that reduces attention noise through differential denoising by splitting each attention head into two, and subtracting one attention map from the other. This mechanism is defined by:

\vspace{-5pt}
{\small
\begin{equation}\label{eq:diffattn}
\text{DiffAttn}(Q_1, K_1, Q_2, K_2, V) = (\alpha_1 - \lambda \alpha_2) V 
\end{equation}}
{\small
\begin{equation}
\alpha_i = \text{softmax}\left(\frac{Q_i {K_i}^T}{\sqrt{d_k}}\right)
\end{equation}
}
%
%

where $\lambda$ is a learnable scalar. To better improve the training dynamics, $\lambda$ is re-parameterized and Group Normalization \citep{wu2018groupnormalization} is applied at the end of each head (post-subtraction).

\subsection{State-Space Layers}
State-space layers were first introduced in \citet{gu2021combining} and were later substantially improved by the S4 model~\citep{gu2022efficiently}. Since then, they have demonstrated strong performance across a variety of domains, including NLP~\citep{fu2022hungry,gss}, audio generation~\citep{goel2022s}, image modeling \citep{baron20232, nguyen2022s4nd}, long-horizon video understanding \citep{wang2023selective}, reinforcement learning \citep{david2022decision,lu2024structured}, and speech recognition \citep{saon2023diagonal}. These models implement linear recurrent update rules derived from time-invariant state-space formulations, which can be efficiently computed in parallel using convolutions and with sub-quadratic complexity.

\subsection{Mamba and Selective State-Space Layers}

A Mamba block processes a signal $U \in \mathbb{R}^{L \times D}$ where D is the hidden dimension and L is the number of tokens. Its core mechanism is the S6 layer, and it forward path formalized by:

\vspace{-10pt}
{\small
\begin{equation}
\label{eq:mamba1}
\begin{split}
X &= \sigma(\text{Conv1D}(\text{Linear}(U))), \quad Z = \sigma(\text{Linear}(U)) \\
Y &= \text{S6}(X), \quad \hat{Y} = \text{Linear}(Y \otimes Z)
\end{split}
\end{equation}
}
where $X$ is the input to the S6 layer, and  to the S6 layers, $X,Z,Y,\hat{Y} \in \mathbb{R}^{L \times D}$. The function $\sigma$ represents SiLU activation, and $\otimes$ represents element-wise multiplication with the gating branch. Each Mamba block is primarily parameterized by linear and convolutional layers, along with the internal components of the S6 layer described below.
\paragraph{S6}
The S6 layer is the most popular variant of SSMs, and it employ real, diagonal and selective SSM. Standard real and diagonal SSMs parameterized by a diagonal transition matrix $A \in \mathbb{R}^{N' \times N'}$, input and output matrices $B,C \in \mathbb{R}^{N' \times 1}$ where $N'$ is the state size, and a timescale $\Delta \in \mathbb{R}$. Each channel of such an SSM can be viewed as a mapping from an input scalar sequence $x$ to an output scalar sequence $y$ via the following recurrent rule:
%

\vspace{-10pt}
{\small
\begin{equation}
\label{eq:recRule}
\begin{split}
h_t &= \bar{A} h_{t-1} + \bar{B}x_t, \quad y_k = C h_t \\
\bar{A} &= f_A (A, \Delta), \quad \bar{B} = f_B (B, \Delta)
\end{split}
\end{equation}
}
where $f_A,f_B$ are discretization functions, and the discrete system matrices are $\bar{A} \in \mathbb{R}^{N' \times N}$ and $\bar{B} \in \mathbb{R}^{N' \times 1}$. The recurrent rule in Eq.~\ref{eq:recRule} can be computed efficiently in parallel on modern hardware accelerators using work-efficient parallel scans~\citep{smith2022simplified} or a simple scalar convolution via FFTs~\citep{gu2021combining}. Note that Eq.~\ref{eq:recRule} is a map from $\mathbb{R}^L$ to $\mathbb{R}^L$, and to process $D$ channels, multiple independent instances are used.

The S6 layer differs from standard SSMs by employing a selective mechanism, where the system matrices are input-dependent. As a result, the system becomes time-invariant, with the per-step system matrices determined by the entire set of channels and then applied to process each channel independently. The entire mechanism can be computed by: $S_B, S_C \in \mathbb{R}^{N' \times D}$, $A \in \mathbb{R}^{D \times N'}$ and $S_{\Delta} \in \mathbb{R}^{1 \times D}$ to define the time-variant matrices by:
%
%

\vspace{-10pt}
{\small
\begin{equation}
\label{eq:TimeVariantMatrices1}
\begin{split}
B_t &= S_B X_{*t}, \; C_t = S_C X_{*t}, \; \Delta_t = \text{softplus}(S_{\Delta} X_{*t}) \\
\bar{A}_t &= \exp(\Delta_t A), \; \bar{B}_t = \Delta_t B_t
\end{split}
\end{equation}
}

and the time-variant recurrent rule by:
{\small
\begin{equation}\label{eq:timeVaraintRecRule}
     h_t =  \bar{A}_t h_{t-1} + \bar{B}_t x_t, \quad y_k = C_t h_t
\end{equation}
}
We study the 'many-to-one' setting, where the model processes an entire input sequence to produce a single output. This regime is widely used in NLP, encompassing both auto-regressive next-token prediction and sequence classification tasks.

\subsubsection{Mamba as Implicit Attention}
The connection between Mamba and linear attention layers is well-established in~\citep{ali2024hidden,dao2024transformers,sieber2025understanding}. Specifically, it has been shown that the time-variant recurrent update rule of S6 for a single channel (see Eq.~\ref{eq:timeVaraintRecRule}) can be explicitly unrolled into the linear attention formulation $Y=AX$ when $A$ is an implicit attention matrix defined by:

\vspace{-10pt}
\begingroup
\setlength{\arraycolsep}{2pt} 
\small
\begin{equation}\label{eq:S6asAttn}
\begin{bmatrix}
    C_{1}\bar{B}_{1} & 0 & \cdots & 0 \\
    C_{2}\bar{A}_{2}\bar{B}_{1} & C_{2}\bar{B}_{2} & \cdots & 0 \\
    \vdots & \vdots & \ddots & 0 \\
    C_{L}\!\!\prod_{k=2}^L \bar{A}_{k}\bar{B}_{1} &
    C_{L}\!\!\prod_{k=3}^L \bar{A}_{k}\bar{B}_{2} &
    \cdots &
    C_{L}\bar{B}_{L}
\end{bmatrix}
\end{equation}
\endgroup

This perspective is further extended by~\citet{zimerman2024unified}, who generalize the interpretation of S6 as implicit attention from Eq.~\ref{eq:S6asAttn} to encompass most components of the Mamba block, including activations, normalization layers, convolutional layers, and the gate branch, into a unified implicit attention formulation. 

\paragraph{Data-Controlled Linear Operators}
The formulation of data-controlled linear operators was first introduced by~\citet{poli2023hyena}, who demonstrated that self-attention can be viewed as an expressive form of such operators. This principle guided the authors in designing the Hyena layer. Subsequently,~\citet{ali2024hidden} showed that S6 layers could also be unified under an implicit variant of this formulation, which~\citet{zimerman2024unified} further extended to additional architectures, including the entire Mamba block, RWKV~\citep{peng2023rwkv}, and Griffin~\citep{de2024griffin}. This extended perspective inspired our approach, leading us to interpret differential design as a method to implicitly parameterize less noisy data-controlled linear operators. This insight motivated our decision to apply differential design at the Mamba level.

\section{Method}

\label{sec:method}
We begin with the simplest implementation of incorporating the differential mechanism into the Mamba block. Our approach is inspired by the Differential Transformer~\citep{ye2024differential}, which applies subtraction at the attention-level rather than at the transformer level. This mechanism is built on top of self attention and it can be described as:

\vspace{-10pt}
{\small
\begin{equation}
    \forall i \in [1,2]: Q_i = X W^Q_i, K_i = X W^K_i,\quad V = X W^V 
\end{equation}
}
{\small
\begin{equation}\label{eq:diffattention}
 \text{DiffAttn}(x) = \Big{(} \mathcal{S}(\frac{Q_1 K_1^T}{\sqrt{d}}) - \lambda  \mathcal{S}(\frac{Q_2 K_2^T}{\sqrt{d}}) \Big{)} V \,
\end{equation}
}
Here, $\mathcal{S}$ is the softmax 
and $\lambda$ is 
parameterized to ensure stability and improve training dynamics:

\vspace{-10pt}
{\small
\begin{equation}\label{eq:lambda}
 \lambda = \exp(\lambda_{q1} \cdot \lambda_{k1}) -\exp (\lambda_{q2} \cdot \lambda_{k2}) - 
 \lambda_{\text{init}}   
\end{equation}
}
To better adapt this technique to Mamba models, we reinterpret differential attention through the lens of a data-controlled linear operator~\citep{poli2023hyena}, leading to the following formulation:

\vspace{-10pt}
{\small
\begin{equation}
\label{eq:diffAttnDataCtrl}
\begin{split}
\text{DiffAttn}(x) &= AV, \quad A = A_1 - \lambda A_2 \\
A_1 &= \mathcal{S}\left(\frac{Q_1 K_1^T}{\sqrt{d}}\right), \quad A_2 = \mathcal{S}\left(\frac{Q_2 K_2^T}{\sqrt{d}}\right)
\end{split}
\end{equation}
}
here, the matrix A defines a data-controlled linear operator.

\begin{figure*}[t]
    \centering
    \includegraphics[width=1.0\linewidth]{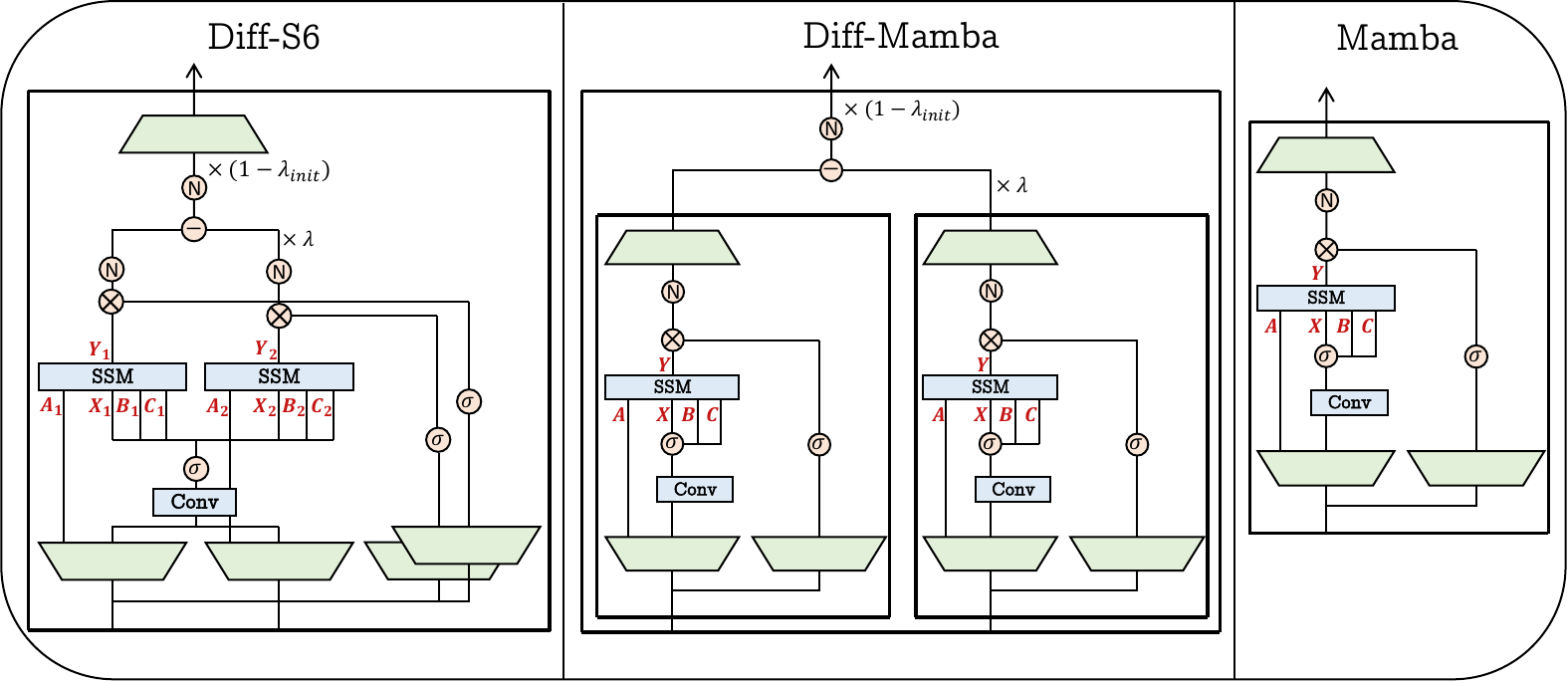}
     \vspace{-6pt}
    \caption{Comparative illustration of our variants Diff-Mamba and Diff-S6 versus the original Mamba architecture, where $\otimes$ is elementwise multiplication, $\sigma$ is the SILU activation, Linear and Conv1D are standard linear projection and 1-dimensional convolution layers, and N stands for normalizations.}
    \label{fig:DiffVairants}
    \vspace{-4pt}
\end{figure*}

\subsection{Diff S6}
Eq.~\ref{eq:diffAttnDataCtrl} defines a straightforward approach to implementing differential Mamba by subtracting values obtained from S6 layers instead of attention layers. This builds upon two key similarities between S6 and attention: (i) S6 in Mamba serves the same role as attention in Transformers--capturing interactions between tokens, and (ii) S6 layers have been shown to be an implicit form of causal linear attention.

Thus, incorporating the differential mechanism into the Mamba block can be achieved by:

{\small
\begin{equation}\label{eq:diffs6}
    \text{Diff S6}(X) = {S6}_1 (X) - \lambda {S6}_2 (X)
\end{equation}
}
where $\lambda$ is defined similarly to Eq.~\ref{eq:lambda}.

Similar to Eq.~\ref{eq:diffAttnDataCtrl}, this formulation can be rewritten in the form of a data-controlled linear operator:

{\small
\begin{equation}\label{eq:diffs6DataCtrl}
    \text{Diff S6}(x) = (A_1 -  \lambda A_2) X = AX
\end{equation}
}
where $A_1$ and $A_2$ are the implicit attention matrices of S6 controlled by the system matrices $\bar{A}_{ij}$,$\bar{B}_{ij}$ and $C_{ij}$ for any time-step $j\in [L]$ and model index $j \in {1,2}$ defined as follows:

\vspace{-10pt}
\begingroup
\setlength{\arraycolsep}{2pt} 
\small
\begin{equation}
\begin{bmatrix}
    C_{i1}\bar{B}_{i1} & 0 & \cdots & 0 \\
    C_{i2}\bar{A}_{i2}\bar{B}_{i1} & C_{i2}\bar{B}_{i2} & \cdots & 0 \\
    \vdots & \vdots & \ddots & 0 \\
    C_{iL}\!\!\prod_{k=2}^L \bar{A}_{ik}\bar{B}_{i1} & 
    C_{iL}\!\!\prod_{k=3}^L \bar{A}_{ik}\bar{B}_{i2} & 
    \cdots & 
    C_{iL}\bar{B}_{iL}
\end{bmatrix}
\end{equation}
\endgroup
One crucial difference between Diff S6 and Diff Attention (see Eqs.~\ref{eq:diffs6} and  ~\ref{eq:diffattention}) is that Diff Attention subtracts elements on the same scale, as softmax produces values in the range [0,1]. In contrast, S6 produces unnormalized and unbounded outputs. To address this discrepancy, we introduce an additional normalization step denoted by $\mathbb{N}$:

{\small
\begin{equation}\label{eq:normdiffS6}
    \mathbb{N}\text{-Diff S6}(X) = \mathbb{N}({S6}_1 (X) - \lambda{S6}_2 (X))
\end{equation}
}
For simplicity, we define $\lambda$ as:

{\small
\begin{equation}
    {\lambda} = \text{Sigmoid}{( \sum\bar{\lambda})} +  \lambda_{\text{init}}
\end{equation}
}
where $\bar{\lambda} \in \mathbb{R}^D$ is a learnable parameter used to parameterize $\lambda$ as a positive and more stable weight.

\subsection{Diff-Mamba}
However, as detailed in the results section, Diff S6 does not perform well and, in practice, falls short of standard Mamba layers. We suspect this arises from S6 being too simple and not functioning as a general-purpose, expressive mixing alternative to attention layers. Consequently, it fails to leverage the full potential of differential techniques.
To address this limitation, we draw inspiration from \citet{zimerman2024unified}, who demonstrated that the entire Mamba block can function as an alternative mixing mechanism to attention by formulating it as a data-controlled linear operator. In particular, the authors show that Mamba can be reformulated as implicit attention by:
%
%

\vspace{-3pt}
{\small
\begin{equation}
\text{Mamba}(X) = AX
\end{equation}
\begin{equation}\nonumber
A = \text{SILU}(\text{Linear}(x)) \hat{\alpha}
 \text{ diag } (\text{Sig}(\text{Conv}(x))) M
\end{equation}
}
where M is a matrix representing the convolution layer, and $\hat{\alpha} $ is the linear operator corresponding to S6, as formalized by~\citet{ali2024hidden}.
This formulation characterizes Mamba as a data-control linear operator with richer and more expressive implicit attention matrices. Building on this insight, we introduce Diff-Mamba, a mechanism that extends the differential approach to the full Mamba block:

\vspace{-8pt}
{\small
\begin{equation}
    \text{Diff-Mamba}(X) = {\text{Mamba}}_1 (X) - \lambda {\text{ Mamba}}_2 (X)
\end{equation}
}
Similar to Eqs.~\ref{eq:diffAttnDataCtrl} and ~\ref{eq:diffs6DataCtrl}, this can be rewritten as a data-controlled linear operator, defined by:

{\small
\begin{equation}
    \text{Diff-Mamba}(X) = AX, \quad A = A_1 - \lambda A_2
\end{equation}
}

A key distinction between Diff Attention and Diff-Mamba is that the latter applies subtraction across a broader set of components, as illustrated in Figure~\ref{fig:DiffVairants}.


Similar to the normalized Diff S6 variant in Eq.~\ref{eq:normdiffS6}, we add a normalization term:

{\small
\vspace{-12pt}
\begin{equation}\label{eq:normalizedDiffMamba}
\vspace{-3pt}
 \mathbb{N}\text{-Diff-Mamba}(x) = \mathbb{N} (\text{Mamba}_1 (x) - \lambda \text{Mamba}_2 (x))
\end{equation}
}
Resulting in the normalized Diff-Mamba mechanism, which is our primary contribution. 
Finally, following~\citet{ye2024differential}, we multiply the output of all variants by $1-\lambda_{\text{init}}$.



\section{Experiments}
\label{sec:experiments}

In this section, we empirically evaluate the effectiveness of the Diff-Mamba architecture. We begin in Section~\ref{subsec:resLM} by demonstrating that Diff-Mamba outperforms the original Mamba architecture in small-scale language modeling tasks across multiple datasets. In Section~\ref{subsec:resAblations}, we justify our key design decisions through a comprehensive series of ablation studies. Subsequently, in Section~\ref{sec:scaling}, we use carefully designed synthetic tasks that are predictive of model behavior in large scale settings. Then, in Section~\ref{sec:medium_scaling} we train the models at medium scale and test long-context capabilities. Afterwards, in Section~\ref{subsec:resRetrieval} we showcase the superior retrieval performance of Diff-Mamba relative to Mamba. Finally, in Section~\ref{subsec:resLens}, we utilize tools from the domain of mechanistic interpretability, such as Tuned-lens \citep{belrose2023eliciting}, to examine the internal representations of Diff-Mamba in comparison to Mamba, empirically validating that our differential approach effectively reduces noise in intermediate representations. A full description of our experimental setup, as well as an efficiency analysis, is provided in Appendix~\ref{sec:app_Expsetup} and Appendix~\ref{sec:efficiency}.

\begin{table}[t]
    \centering
    \small
    \begin{tabular}{@{\hspace{3pt}}l@{\hspace{6pt}}c@{\hspace{6pt}}c@{\hspace{6pt}}c@{\hspace{6pt}}c@{\hspace{3pt}}}
        \toprule
        Model & Dataset & \# Layers & \# Params & PPL~$(\downarrow$) \\
        \midrule
        Mamba       & Wikitext-103 & 6  & 167M  & 22.357 \\
        Diff-Mamba  & Wikitext-103 & 6  & 169M  & \textbf{22.282} \\
        \midrule
        Mamba       & Wikitext-103 & 12 & 255M   & 20.413     \\
        Diff-Mamba  & Wikitext-103 & 12 & 259M   & \textbf{20.012}    \\
        \midrule
        Mamba       & Text8        & 6  & 127M  & 2.416  \\
        Diff-Mamba  & Text8        & 6  & 129M  & \textbf{2.396}  \\
        \midrule
        Mamba       & Text8        & 12 & 255M  & 2.525  \\
        Diff-Mamba  & Text8        & 12 & 259M  & \textbf{2.479}  \\
        \midrule
        Mamba       & Enwik8       & 6  & 127M  & 2.321  \\
        Diff-Mamba  & Enwik8       & 6  & 129M  & \textbf{2.314}  \\
        \midrule
        Mamba       & Enwik8       & 12 & 255M  & 2.422  \\
        Diff-Mamba  & Enwik8       & 12 & 259M  & \textbf{2.381}  \\
        \bottomrule
    \end{tabular}
    \caption{Final performance of Mamba and Diff-Mamba across model sizes. All models were trained for 40 epochs on each dataset. Trends shown in Figure~\ref{fig:trainingCurves}.}
    \vspace{-4pt}
    \label{tab:final_res_lm}
\end{table}

\begin{figure*}[!h]
  \centering
\begin{tabular}{@{}c@{\hspace{0.5em}}c@{\hspace{0.5em}}c@{}}
  \includegraphics[width=0.324\linewidth]{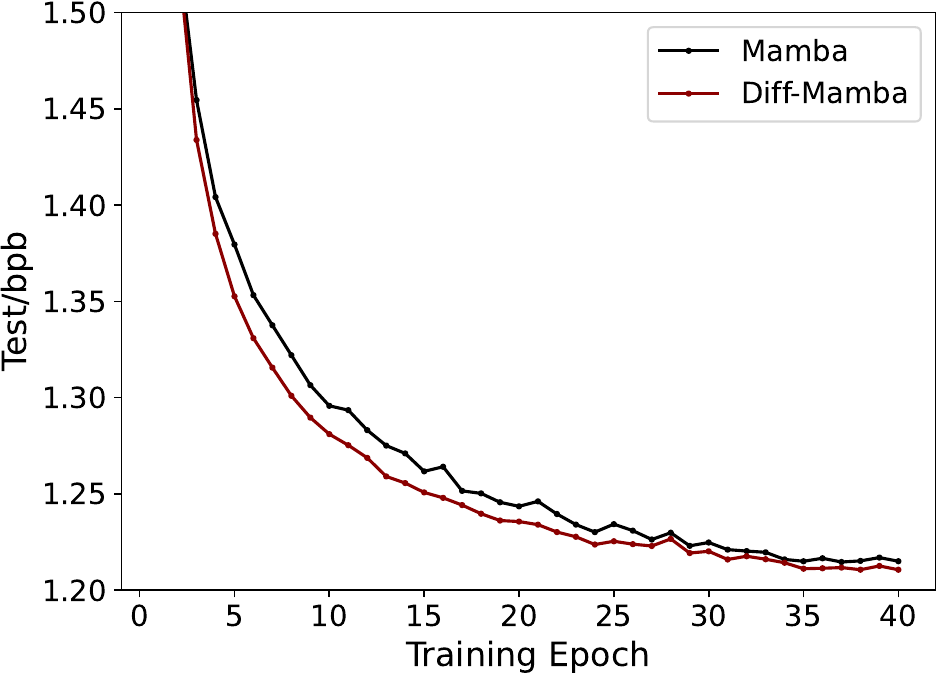} &
\includegraphics[width=0.324\linewidth]{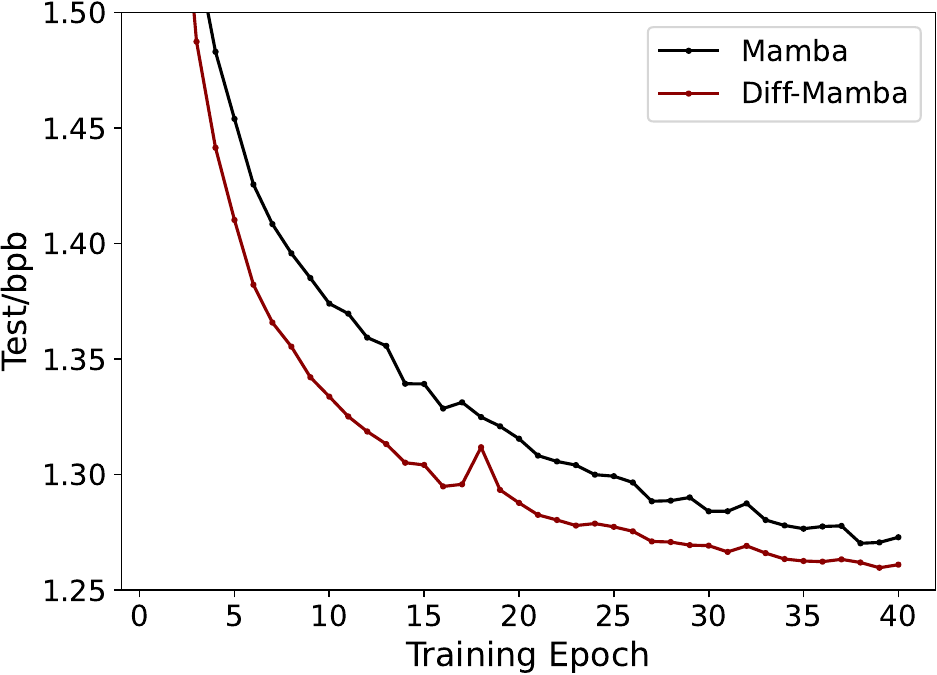} &
\includegraphics[width=0.324\linewidth]{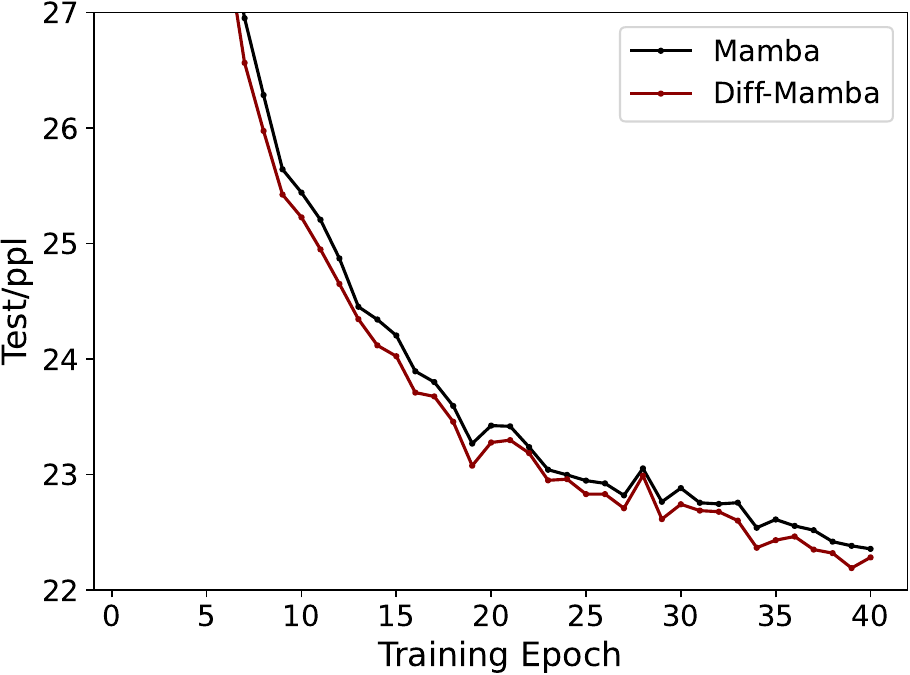} \\
\includegraphics[width=0.324\linewidth]{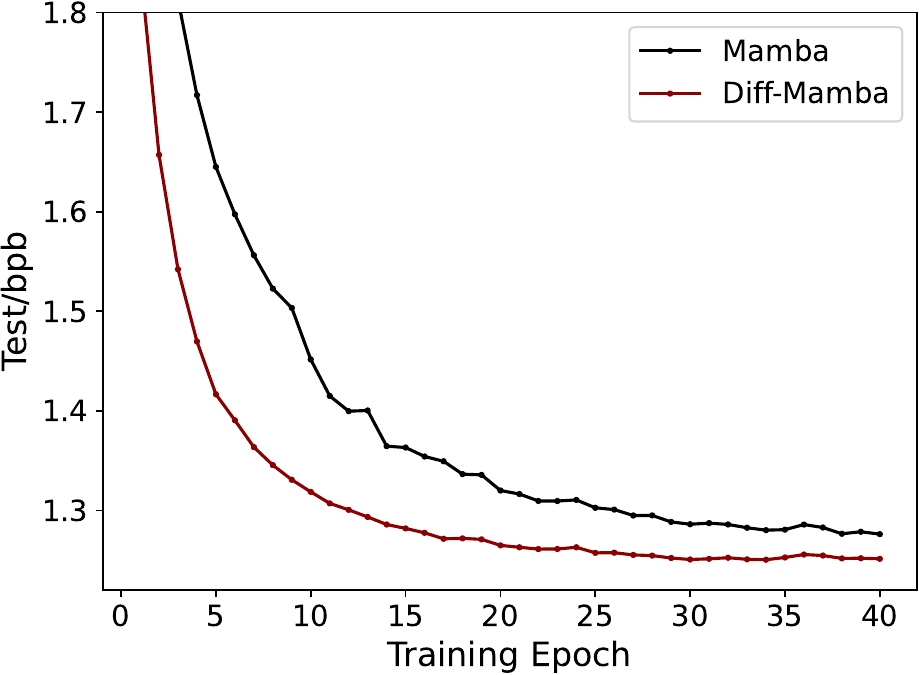} &
\includegraphics[width=0.324\linewidth]{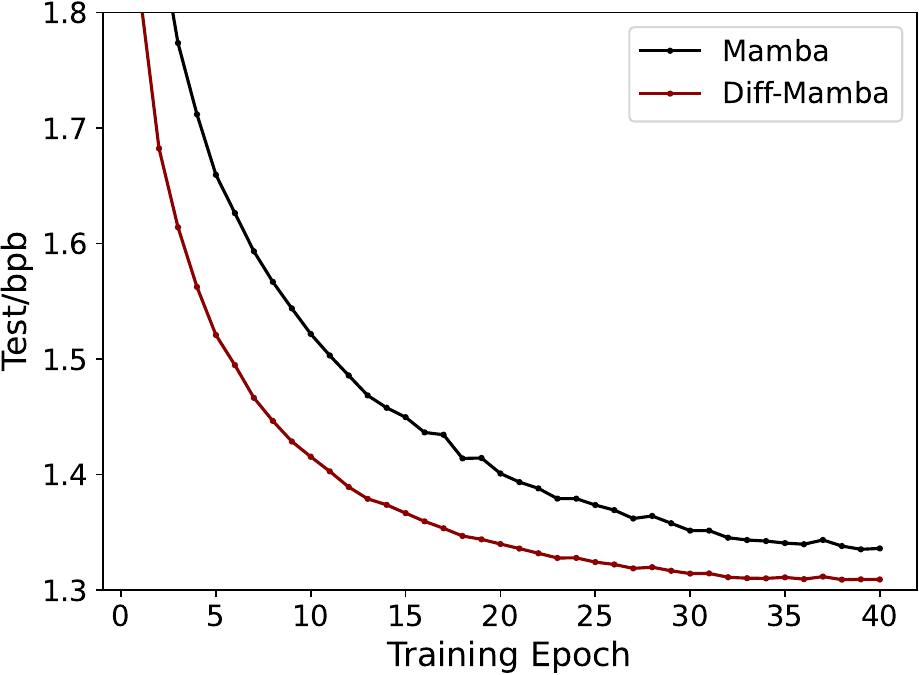} &    
\includegraphics[width=0.324\linewidth]{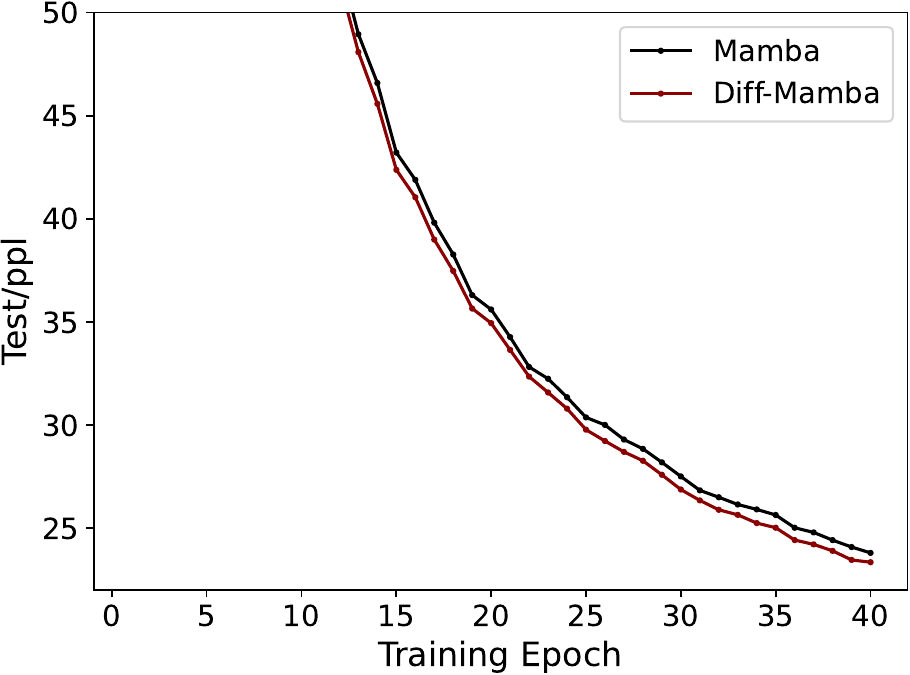} 
  \end{tabular}
  \vspace{-5pt}
  \caption{Comparison of test curves through the training for Mamba and Diff-Mamba. The top row shows results for 6-layer models, and the bottom row for 12-layer models. Columns correspond to datasets: Enwik8 (left), Text-8 (center), and WikiText-103 (right). 
  }
  \label{fig:trainingCurves}
  \vspace{-2pt}
\end{figure*}

\vspace{-3pt}
\subsection{Language Modeling}\label{subsec:resLM}
To evaluate the performance of Diff-Mamba relative to Mamba on general NLP tasks, we train both models from scratch using comparable model sizes and an identical training setup, including the same codebase \citep{gu2022efficiently}, datasets, and hyperparameters. We focus on three widely used benchmarks: WikiText-103, Text8, and Enwik8, and experiment with models of varying depth. The final results are reported in Table~\ref{tab:final_res_lm}. To provide a more comprehensive view of the optimization process, we include test curves through the training in Figure~\ref{fig:trainingCurves}.

As shown in Table~\ref{tab:final_res_lm}, Diff-Mamba outperforms Mamba across all evaluated environments, achieving consistently lower loss. In particular, for the 12 layer model, Diff-Mamba improves over Mamba by 0.4 perplexity on WikiText-103, 0.046 bits per byte (bpb) on Text8, and 0.041 bpb on Enwik8. For 6 layer model, Diff-Mamba improves over Mamba by 0.075 on WikiText-103, 0.02 on Text8, and 0.007 on Enwik8. Interestingly, we observe that as the number of layers increases, the differential design in Mamba becomes increasingly effective. A possible explanation for this is that, in the lower layers, the dependencies captured by the implicit attention matrices are shorter and simpler. In these cases, Mamba can manage overallocation effectively without the need for a differential mechanism. However, in the upper layers, the dependencies become more complex, spanning longer ranges~\citep{ben-kish2025decimamba} and exhibiting more diverse patterns. This amplifies the impact of overallocation, thereby making the benefits of the differential design more pronounced. Furthermore, the training curves in Figure~\ref{fig:trainingCurves} provide insight into the optimization properties of Diff-Mamba, showing that it consistently outperforms Mamba and achieves faster convergence. We hypothesize that this phenomenon arises from the fact that the differential design reduces the amount of noise, which appears to be critical for improving convergence~\citep{NIPS2013_ac1dd209,zhang2019lookahead}.

\subsection{Ablations Analysis}\label{subsec:resAblations}

\begin{table}[t]
    \small
    \centering
        \begin{tabular}{@{\extracolsep{\fill}}lccc}
        \toprule
        Model & w.o Nrm & w. Nrm & \# Params \\
        \midrule
        Mamba & 2.577 & -- & 127M  \\
        Diff-S6  & 2.520 & 2.512 & 128M\\
        Diff-S6 + re. $\bar\lambda$ & 2.529 & 2.517 & 128M\\
        Diff-Mamba  & 2.508 & \textbf{2.493} & 128M \\
        Diff-Mamba + re. $\bar\lambda$ & 2.517 & 2.503 & 128M \\
        \bottomrule
    \end{tabular}
    \caption{\textbf{Ablation study} comparing (i) the scope at which the differential mechanism is applied (Diff-S6 vs. Diff-Mamba), (ii) the effect of including normalization ("w. Nrm") versus excluding it ("w.o Nrm"), and (iii) the importance of reparameterization for $\bar\lambda$ ("re. $\bar\lambda$"). All models were trained on full Text8 with an identical parameter count. Reported values are test perplexity (PPL) on epoch 10. Lower is Better.}
    \label{tab:ablations}
    \vspace{-4pt}
\end{table}

To validate our design decisions regarding (i) applying the differential operation at the S6 versus Mamba layer, (ii) incorporating an additional normalization sub-layer before subtraction, and (iii) reparameterization $\bar\lambda \in \mathbb{R}^D$ to a scalar, we conducted dedicated ablation experiments on the Text8 benchmark. Table~\ref{tab:ablations} summarizes the results. All models share an identical parameter count and were trained with same hyper-parameters that optimized for the baseline 
model. 
It can be seen that all three design choices are justified, specifically Diff-S6 outperforms Mamba without normalization, while Diff-Mamba outperforms Diff-S6. Incorporating additional normalization leads to improved results, yielding gains of 0.015 and 0.008 perplexity for Diff-Mamba and Diff-S6, respectively. Finally, $\bar\lambda$ reparameterization doesn't contribute to better performance demonstrated both in Diff-S6 and Diff-Mamba models.


\subsection{Unit Tests as Predictors of Scaling}
\label{sec:scaling}
\begin{figure*}
    \centering
    \includegraphics[width=\linewidth]{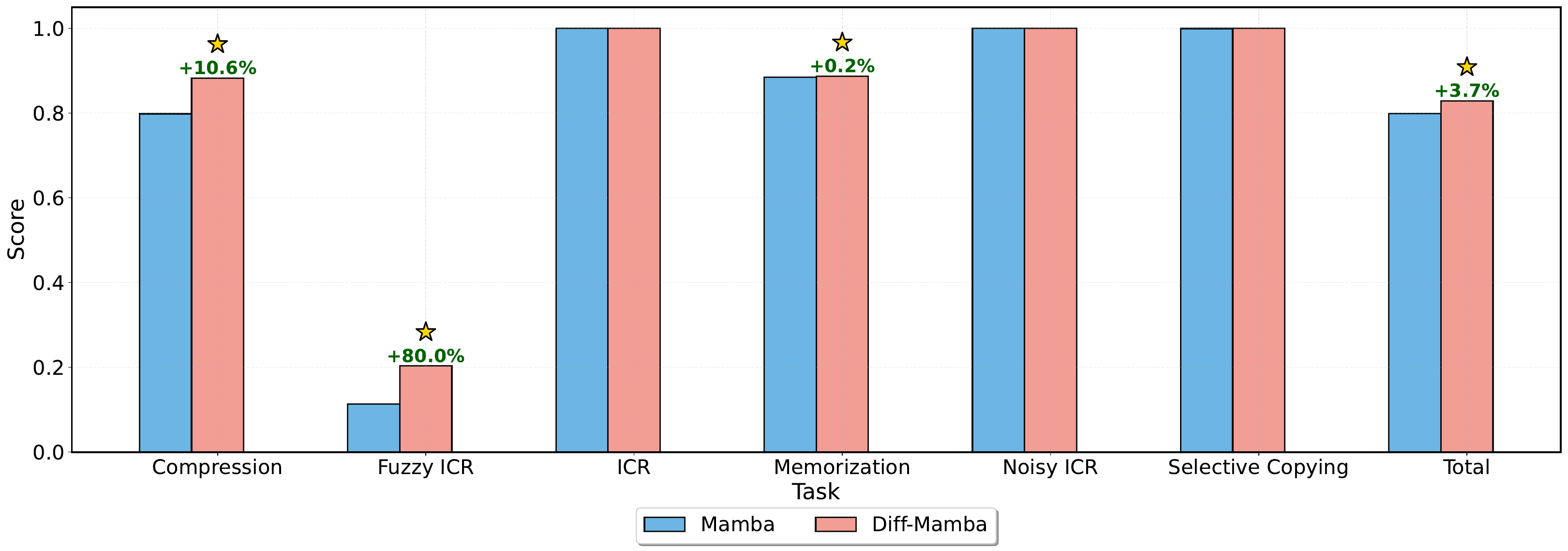}
    \vspace{-15pt}
    \caption{Performance comparison on synthetic token manipulation tasks. We evaluate Mamba and Diff-Mamba architectures across six synthetic capability benchmarks. Diff-Mamba consistently outperforms Mamba, with notable improvements in Fuzzy ICR of 80.0\% and Compression with 10.6\%. Stars indicate the best model per task.}
    \label{fig:atomic_tasks}
\end{figure*}

Training LLMs at scale is prohibitively resource-intensive. Nevertheless, prior work has established that carefully designed small-scale capability evaluations \cite{gupta2022simplifying} are predictive of model behavior in large-scale \cite{poli2024mechanistic}. To this end, we follow the MAD pipeline \cite{poli2024mechanistic} and adopt a rigorous suite of synthetic token manipulation tasks, that serve as capability unit tests. These tasks allow us to isolate and assess core mechanisms that underpin scalable performance. We investigate Mamba and Diff-Mamba. Diff-Mamba model consistently demonstrates superior results (Figure~\ref{fig:atomic_tasks}), providing strong evidence that the architecture is well-positioned to retain and extend these capabilities when trained at larger scales. Both models achieve the highest accuracy on in-context-recall (ICR), noisy ICR, and selective-copying tasks, while Diff-Mamba outperforms Mamba on the rest of the tasks, specifically up to 10.6\% improvement in compression, 80\% in fuzzy ICR, and 0.2\% in memorization. Diff-Mamba surpasses Mamba by 3.7\% in total.

\subsection{Early Results at Medium Scale}
\label{sec:medium_scaling}
To thoroughly evaluate Diff-Mamba, we trained both Mamba and Diff-Mamba from scratch, each with 370M parameters, on a 50B token subset of The Pile dataset~\citep{gao2020pile}.
After an ablation study (Figure \ref{fig:loss_the_pile} in the Appendix), the most effective Diff-Mamba variant for scalable performance was found to be a repeated alternation of Mamba and Diff-Mamba layers throughout the network. This hybrid variant performs better than both models with fully Mamba or Diff-Mamba layers.

\begin{figure}[t]
    \small
    \centering
    \includegraphics[width=0.8\linewidth]{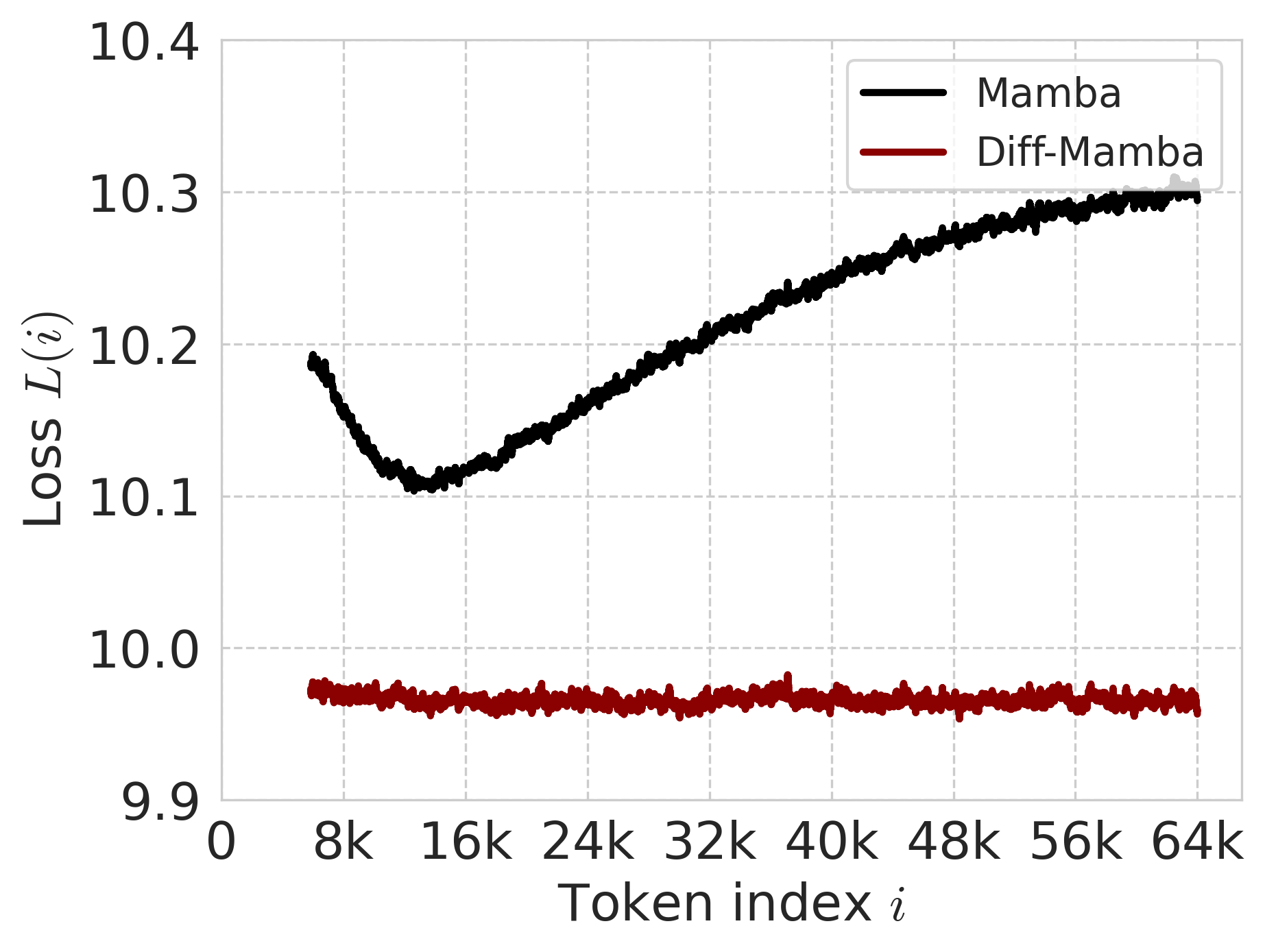}
    \vspace{-8pt}
    \caption*{(a)}
    \vspace{4pt} 
    
    \renewcommand{\arraystretch}{1.2}
    \begin{tabular}{lccc}
        \toprule
        \textbf{Model} & \textbf{The Pile~$(\downarrow$)} & \textbf{PG19~$(\downarrow$)} \\
        \midrule
        Mamba          & 11.212 & 28.064 \\
        Diff-Mamba     & 11.081 & 26.619 \\
        \bottomrule
    \end{tabular}
    \vspace{-6pt}
    \caption*{(b)}
    
    \vspace{-6pt}
    \caption{Diff-Mamba excels in both tests. (a) The x-axis is the token index, and the y-axis is the corresponding per-token loss. (b) PPL results on test set of The Pile and PG19.}
    \label{fig:scaling_tests}
\end{figure}

Next, we evaluated the models via zero-shot tests on the LongCrawl64 dataset~\citep{buckman2024}, a long-sequence subset of RedPajama-v2~\citep{weber2024redpajama} designed particularly for research on long-context. In Figure~\ref{fig:scaling_tests} (a), per-token loss was calculated over the dataset following \citep{lin2025forgetting}. In addition, in Figure~\ref{fig:scaling_tests} (b), we calculated PPL on the test subset of The Pile and PG19. Diff-Mamba demonstrates impressive long context capabilities, maintaining per-token loss of around 9.98 across different context lengths, while Mamba increases significantly as the context grows. Furthermore, Diff-Mamba outperforms Mamba by PPL scores of 0.131, and 1.445 on The Pile and PG19 test sets respectively. We consider the positive results at medium scale promising, suggesting that alternating between Mamba and Diff-Mamba layers yields a more effective, scalable, and robust architecture.

\begin{figure}[!h]
    \centering
    \includegraphics[width=1.0\linewidth]{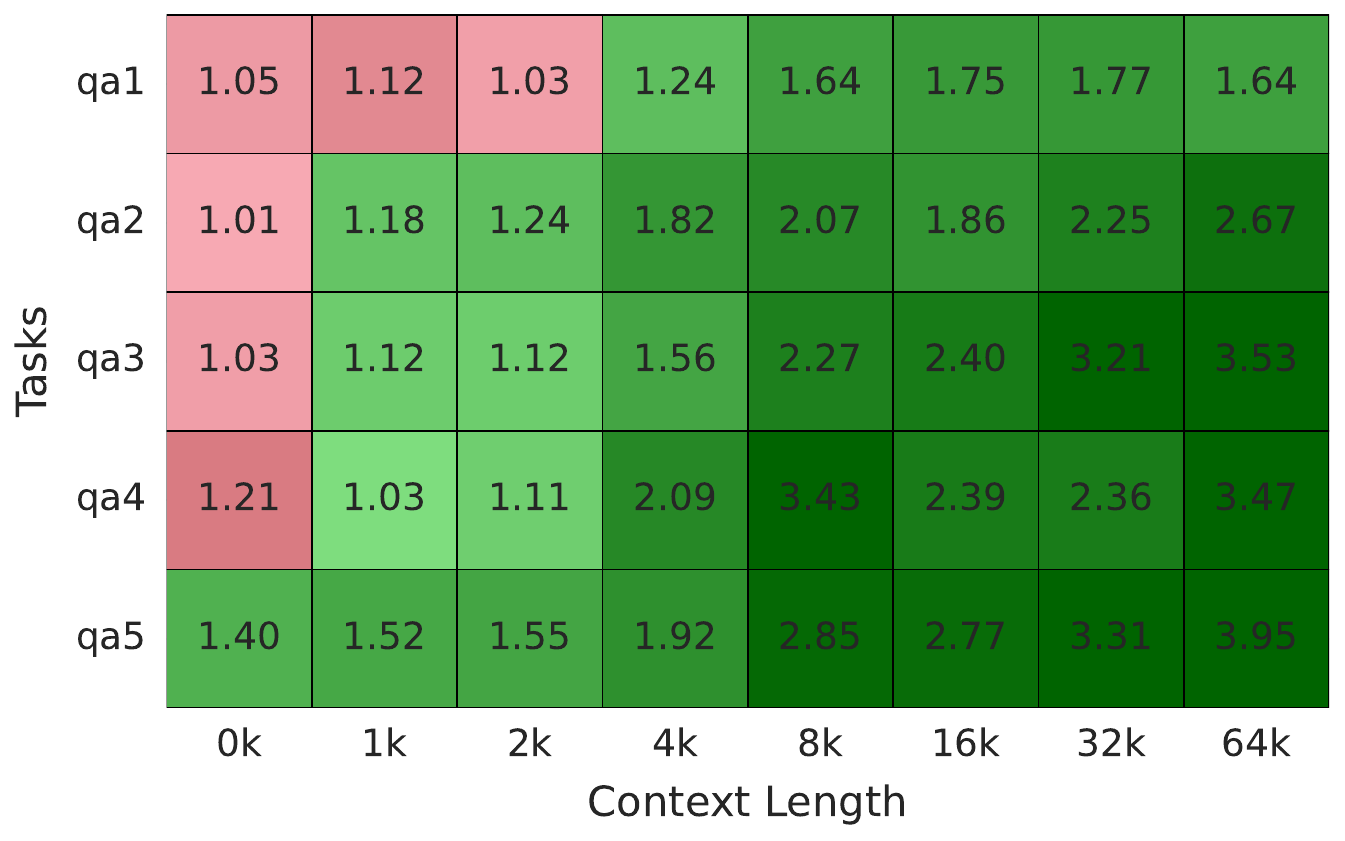}
    \vspace{-17pt}
    \caption{\textbf{Retrieval Abilities: }Comparison of Diff-Mamba and Mamba models across five retrieval tasks from BABILong. X-axis represents the context length, and y-axis corresponds to the task index. Each cell displays the ratio in which one model outperforms the other. Green cells indicate wins by Diff-Mamba, while red cells indicate wins by Mamba.}
    \label{fig:niddle}
    \vspace{-7pt}
\end{figure}

\subsection{Retrieval}\label{subsec:resRetrieval}
The Diff-Transformer exhibits significantly improved retrieval capabilities compared to the original Transformer. Consequently, we conduct experiments across five retrieval tasks from the BABILong benchmark \citep{NEURIPS2024_c0d62e70} to evaluate whether these enhanced abilities transfer to Diff-Mamba. 
BABILong comprises a diverse array of reasoning tasks, including fact chaining, simple induction, and deduction, where relevant facts are embedded within lengthy natural language passages of varying context lengths. This setup provides a rigorous benchmark for evaluating models’ ability to retrieve and reason over extended contexts. To ensure a fair comparison, both Diff-Mamba and Mamba (from Section~\ref{sec:medium_scaling}) were fine-tuned on BABILong tasks with up to 1k tokens to facilitate a more targeted comparison, ensuring that their learning processes were aligned with the same objectives.

Results (Figure~\ref{fig:niddle}) are reported on the test set. Each cell displays the ratio in which one model outperforms the other. Green indicates wins by Diff-Mamba, while red indicates wins by Mamba. As evident from the results, Diff-Mamba outperforms Mamba, exhibiting a slower performance degradation and a larger ratio score as context length increases. Diff-Mamba wins consistently, achieving a ratio of up to 3.95. Our findings demonstrate that Diff-Mamba achieves superior retrieval performance, particularly in long-context scenarios.

\begin{figure*}[!t]
    \centering
    \includegraphics[width=1.0\linewidth]{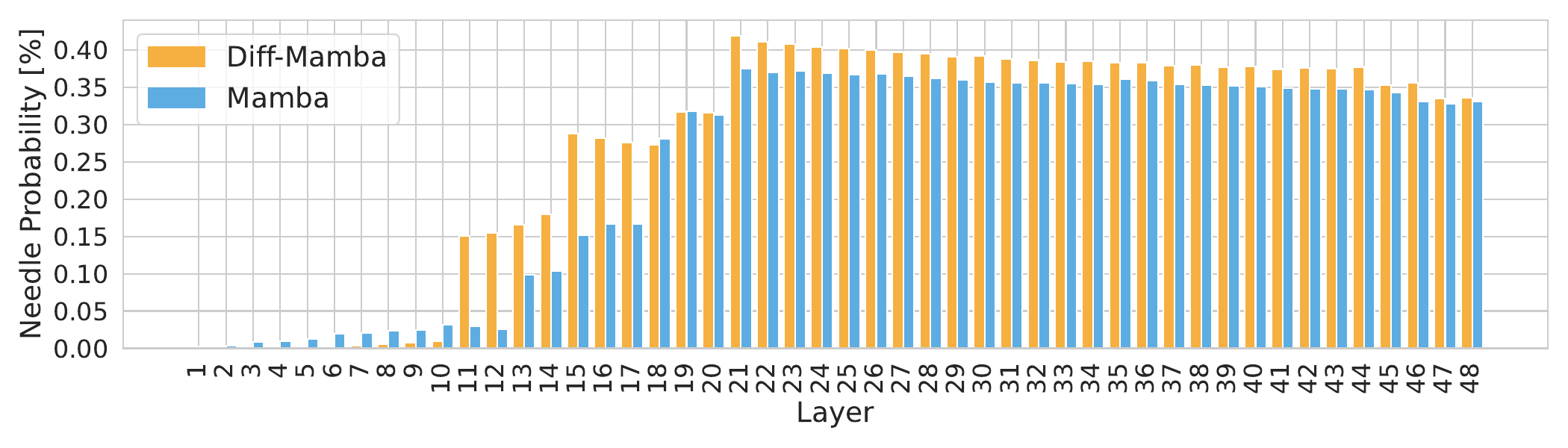}
    \vspace{-20pt}
    \caption{\textbf{Measuring Signal-to-Noise Ratio:} The y-axis represents the probability of predicting the desired needle token, where lower values indicate higher noise. The x-axis denotes various layers within the model where intermediate noise is measured. Results show the average needle probabilities in each layer on 1k examples in BABILong questions of 1k-2k tokens.}
    \label{fig:OverAllocation}\
    \vspace{-13pt}
\end{figure*}

\vspace{-3pt}
\subsection{Noise 
in Intermediate Representations}\label{subsec:resLens}
\vspace{-1pt}
Our empirical analysis in previous sections, as well as the results of Diff-Transformer suggest that the differential design can mitigate the overallocation problem and improve general performance. To further investigate the underlying causes of this phenomenon, we analyze the model's internal representations using tools from the field of mechanistic interpretability. In particular, we leverage Tuned-lens~\citep{belrose2023eliciting} -
a method designed to examine intermediate representations by training an affine probe to map activations at each layer to the model's final prediction, thereby enabling layer-wise interpretability and insight into the model's internal computation. Building on this tool, we measure the signal-to-noise ratio in the hidden representations of Diff-Mamba compared to standard Mamba models. The Tuned-lens tool projects logits from each layer into predictions for the next token on the retrieval task. By measuring the predicted probability of the needle token, we can estimate the signal-to-noise ratio at different layers. Notably, as can be seen in Figure~\ref{fig:OverAllocation}, across the majority of layers Diff-Mamba exhibits a higher signal-to-noise ratio compared to Mamba. This difference is especially pronounced in the early to mid layers, where the predicted probability of the needle token in Diff-Mamba is substantially higher. This analysis empirically demonstrates that Diff-Mamba produces less noisy representations, directly aligning with the key principles underlying the differential mechanism, which is designed to mitigate the overallocation problem.

\section{Discussion: Why 
Differential Design}
A key question that arises from the empirical findings presented in this paper and in Diff-Transformer is why differential design is so effective and what the underlying causes of its power are. Our work sheds light on this question by first showing that the problem of over-allocating attention to irrelevant context is not a phenomenon unique to transformers, but rather a general challenge in architectural research. Moreover, while previous work has primarily motivated the differential design through analogies to noise-canceling headphones and differential amplifiers, and supported it with strong empirical performance, we take this a step further. In Figure~\ref{fig:OverAllocation}, we present a quantitative analysis using tools from the field of model understanding, showing that the intermediate representations in Diff-Mamba exhibit a higher signal-to-noise ratio compared to their non-differential Mamba counterparts, providing empirical support for the motivation previously proposed in the literature. Yet, the underlying reasons behind the empirical success of differential design remain largely unexplored, and further investigation from both theoretical and empirical perspectives is required to advance progress in this important direction.

\section{Conclusions}
\label{sec:conclusion}
In this paper, we introduced Diff-Mamba, a variant of the Mamba architecture that leverages differential design principles to mitigate the problem of attention score over-allocation to irrelevant tokens, thereby enhancing overall performance, with a particular emphasis on long-context, retrieval and robustness capabilities. Diff-Mamba advances architectural research on Mamba variants by introducing an inductive bias that enhances the signal-to-noise ratio, a property that may be critical at scale, as hallucinations remain one of the most significant challenges in LLMs.

\section{Limitations}
\label{sec:Limitations}
Although the experimental results show promising improvements over language modeling and retrieval tasks, we did not provide a rigorous theoretical framework explaining precisely why differential designs improve Transformer or Mamba-based LLMs. Developing such a theoretical justification is left as future work. Additionally, our results are limited to small-to-medium scale experiments due to constraints imposed by our academic budget. Finally, it remains an open question whether differential design principles can be effective in other domains, beyond NLP tasks, for instance, to domains such as computer vision, graph modeling, or time-series analysis.

\section*{Ethics Statement}
Our work aims to improve the robustness of Mamba models by increasing the signal-to-noise ratio through differential design. In the broader context, this contribution supports the development of more efficient, reliable, and trustworthy LLMs. Additionally, our findings highlight when and how differential design is effective and offer practical guidance for incorporating inductive biases that lead to less noisy architectures. Thus, we conclude that our approach contributes to the principled design of safer and more reliable LLMs.

\newpage
\bibliography{references}

\appendix
\newpage

\section{Experimental Setup\label{sec:app_Expsetup}} 
All experiments were conducted with Mamba-2 on an L40s GPU using PyTorch, on publicly available datasets.
\subsection{Language Modeling}
We train all models with a maximum sequence length of 512 for 40 epochs using 3 seeds (0, 42, 77) with settings as in Table~\ref{tab:final_res_lm_conf}. Diff-Mamba uses 1024 channels and 64 heads, while Mamba uses 1024 channels and 128 heads. In Diff-Mamba, we reduce the number of parameters in the linear projections of each Mamba block by half to achieve a similar number of parameters to Mamba. Following~\citep{ye2024differential}, the parameter $\lambda_{\text{init}}$ in Eq. \ref{eq:lambda} is defined as $\lambda_{\text{init}}=0.8-0.6\cdot\exp\!\bigl(-0.3 \,(i_{\text{layer}} - 1)\bigr)$. In addition, the normalization we use in Eq. \ref{eq:normalizedDiffMamba} is RMS-Norm~\citep{zhang2019root}.

\begin{table}[!h]
    \small
    \vspace{10pt}
    \centering
    \begin{tabular}{@{\hspace{3pt}}l@{\hspace{6pt}}c@{\hspace{6pt}}c@{\hspace{6pt}}c@{\hspace{6pt}}c@{\hspace{3pt}}c@{\hspace{6pt}}}
        \toprule
        Model & Dataset & \# Layers & Dropout & Batch & lr \\
        \midrule
        Mamba       & Wikitext-103 & 6  &0.25  & 100 & 5e-4 \\
        Diff-Mamba  & Wikitext-103 & 6  & 0.25  & 100 & 5e-4 \\
        \midrule
        Mamba       & Wikitext-103 & 12 & 0.5  & 50 & 5e-4 \\
        Diff-Mamba  & Wikitext-103 & 12 & 0.5   & 50 & 5e-4 \\
        \midrule
        Mamba       & Text8        & 6  & 0.4  & 170 & 5e-4  \\
        Diff-Mamba  & Text8        & 6  & 0.4  & 170 & 5e-4  \\
        \midrule
        Mamba       & Text8        & 12 & 0.5  & 50 & 5e-5  \\
        Diff-Mamba  & Text8        & 12 & 0.5  & 50 & 5e-5  \\
        \midrule
        Mamba       & Enwik8       & 6  & 0.4  & 170 & 5e-4 \\
        Diff-Mamba  & Enwik8       & 6  & 0.4  & 170 & 5e-4  \\
        \midrule
        Mamba       & Enwik8       & 12 & 0.5  & 80 & 1e-4  \\
        Diff-Mamba  & Enwik8       & 12 & 0.5  & 80 & 1e-4  \\
        \bottomrule
    \end{tabular}
    \caption{Configuration of Mamba and Diff-Mamba across model sizes. All models were trained for 40 epochs on each dataset. Test trends are presented in Figure~\ref{fig:trainingCurves}.\label{tab:final_res_lm_conf}}
\end{table}

\subsection{Ablations Analysis}
To validate our architecture design, an ablation study has been done. To match the number of parameters, each architecture has a different inner parameter division. Additional configuration and models architecture details are presented in Table \ref{tab:hyperparameters_ablations} and Table \ref{tab:ablations_conf}.

\begin{table}[!h]
    \vspace{-5pt}
    \small
   
    \vspace{4pt}
    \centering
    \begin{tabular}{@{\extracolsep{\fill}}lc}
        \toprule
        Params & Values \\
        \midrule
        Layers & 6 \\
        lr & 5e-4 \\
        Warmup Steps  & 1000 \\
        Steps & 15,000 \\
        Droput & 0.4 \\
        Batch Size  & 170 \\
        \bottomrule\\
    \end{tabular}
    \caption{Hyperparameters used for the ablation trainings. \label{tab:hyperparameters_ablations}}
\end{table}

\begin{table}[!h]
    \small
    \centering
    \begin{tabular}{@{\extracolsep{\fill}}lccc}
        \toprule
        Model & \# Channels & \# Heads & Expand \\
        \midrule
        Mamba & 1024 & 128 & 2  \\
        Diff-S6  & 864 & 108 & 2\\
        Diff-S6 + re. $\lambda$ & 864 & 108 & 2\\
        Diff-Mamba  & 1024 & 64 & 1 \\
        Diff-Mamba + re. $\lambda$ & 1024 & 64 & 1 \\
        \bottomrule\\
    \end{tabular}
    \vspace{-10pt}
    \caption{Models architecture details used for the ablation study.\label{tab:ablations_conf}}
\end{table}


\begin{figure*}[!h]
    \centering
    \includegraphics[width=\linewidth]{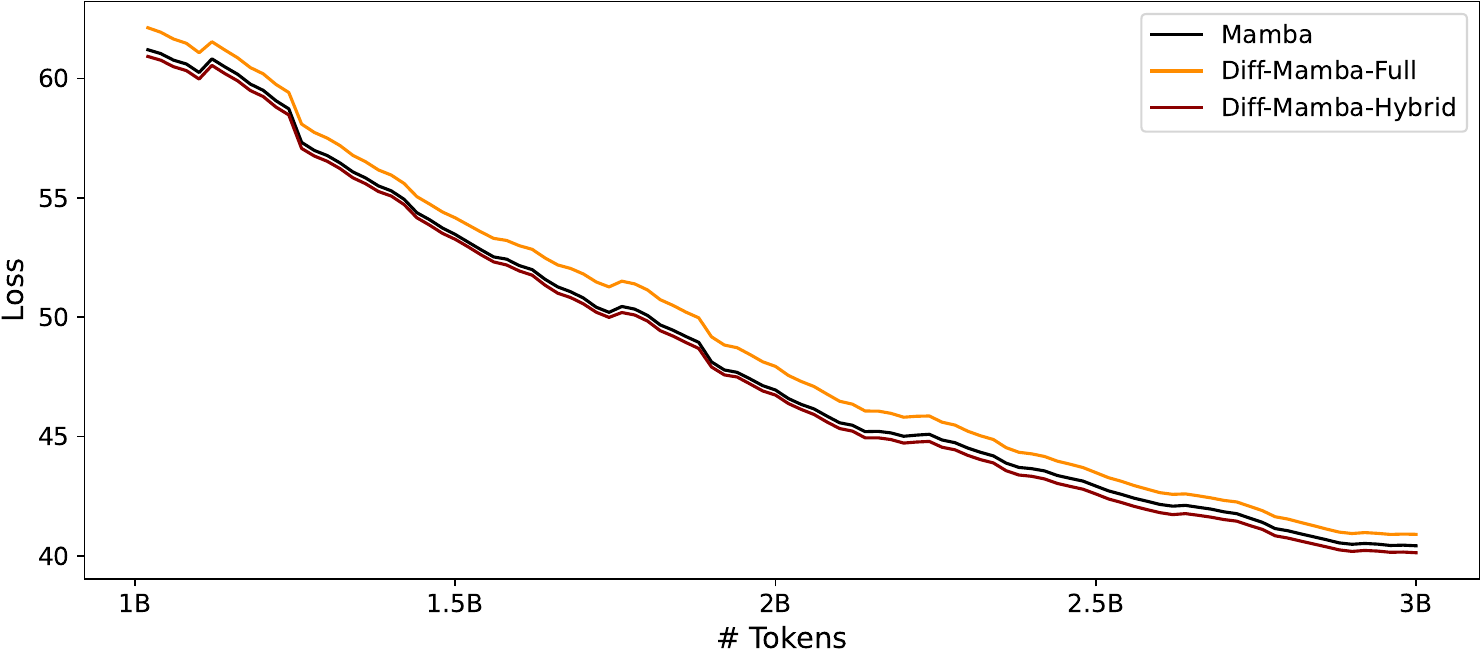}
    \vspace{-18pt}
    \caption{Loss training curve of Diff-Mamba variants compared to Mamba through pre-training on The Pile. Diff-Mamba-Hybrid indicates alternating layers of Mamba and Diff-Mamba, while Diff-Mamba-Full indicates only Diff-Mamba layers model. The trends were smoothed for display adjustment.}
    \label{fig:loss_the_pile}
\end{figure*}

\subsection{Unit Tests as Predictors of Scaling}
\label{sec:scaling_app}
We follow the MAD pipeline \cite{poli2024mechanistic} and adopt a rigorous suite of synthetic token manipulation tasks, that serve as capability unit tests. We investigate Mamba and Diff-Mamba. Each model has four layers, consisting of the core blocks. Hyperparameters and a set of tests were applied similarly to the MAD pipeline, including different numbers of examples for training, learning rates, sequence lengths, and vocabulary size upon tasks such as compression, ICR, and selective-copying.

\subsection{Early Results at Medium Scale}
\label{sec:medium_scalingApp}
We pre-trained both the Mamba and Diff-Mamba architectures and conducted an ablation study on the Diff-Mamba model to evaluate the impact of architectural variations. The results indicate that interleaving Mamba and Diff-Mamba layers throughout the network yeilds improved loss performance during pre-training, compared to using fully Diff-Mamba architecture (Figure~\ref{fig:loss_the_pile}). The pre-training was performed on 8 GPUs over 50 billion tokens using the GPTNeoX tokenizer. The Mamba model contains 368M parameters, Diff-Mamba-Hybrid has 375M parameters, and Diff-Mamba-Full configuration comprises 382M parameters. Although Diff-Mamba-Hybrid has 7 million fewer parameters than Diff-Mamba-Full, it achieves better performance. 
The models were trained with 48 layers, a learning rate of 1.5e-3, 10,000 warm-up steps, the AdamW optimizer with a weight decay of 0.1, a batch size of 1M tokens per step, and a maximum sequence length of 2048 tokens.

\subsection{Retrieval}
For retrieval experiments, we use 50B checkpoint of Diff-Mamba and Mamba trained on The Pile, as described in Section~\ref{sec:medium_scaling} and finetune those on BABILong tasks, followed by an evaluation on a test set. For statistically significant results, we finetuned and evaluated with 3 seeds.

\subsubsection{BABILong Fine-tuning\label{sec:ret_babilongApp}}
\begin{figure}[!h]
    \centering
    \begin{subfigure}[t]{0.495\textwidth}
        \centering
        \includegraphics[width=\linewidth]{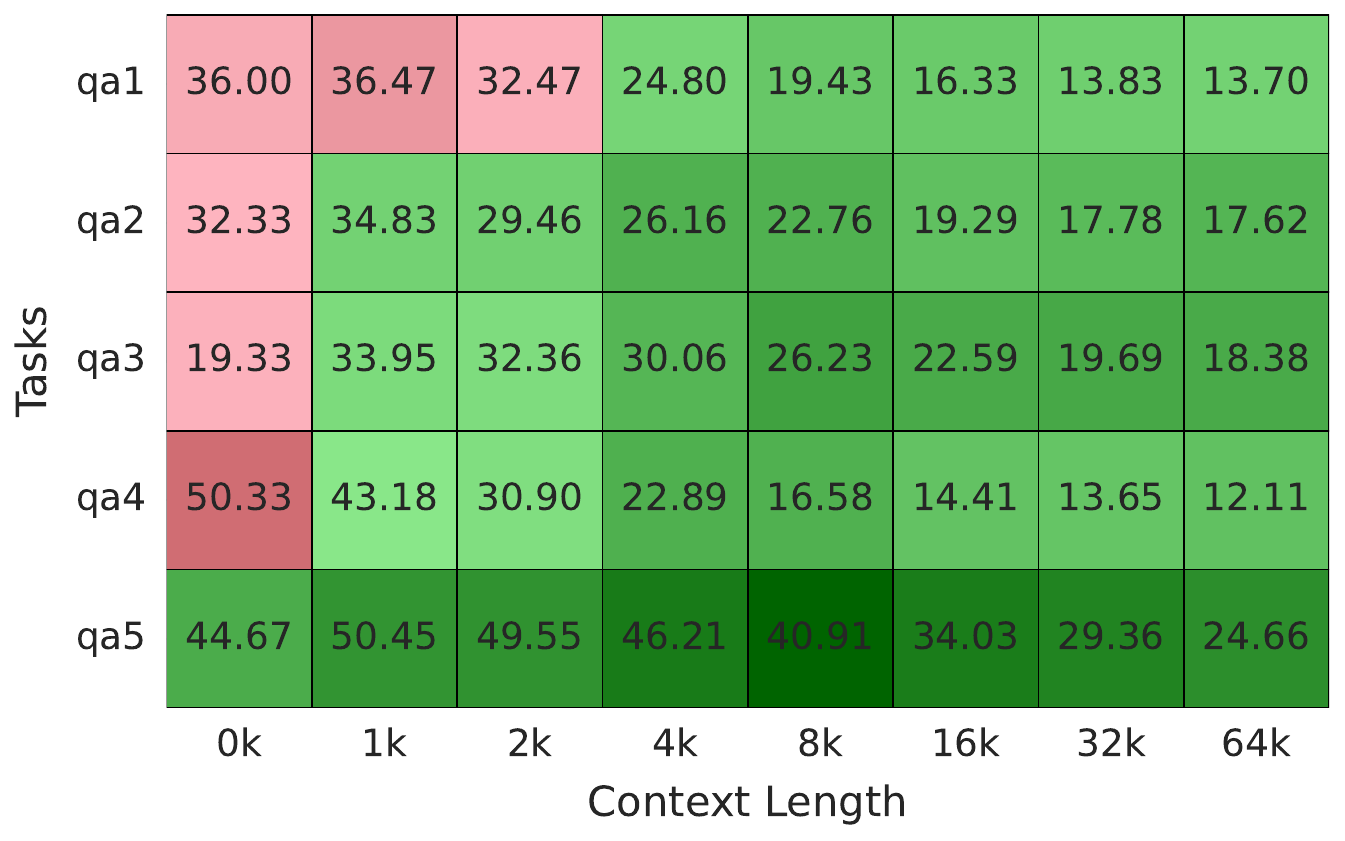}
        \caption{Diff-Mamba}
        \label{fig:subfig1_ft_diffApp}
    \end{subfigure}
    \hfill
    \begin{subfigure}[t]{0.495\textwidth}
        \centering
        \includegraphics[width=\linewidth]{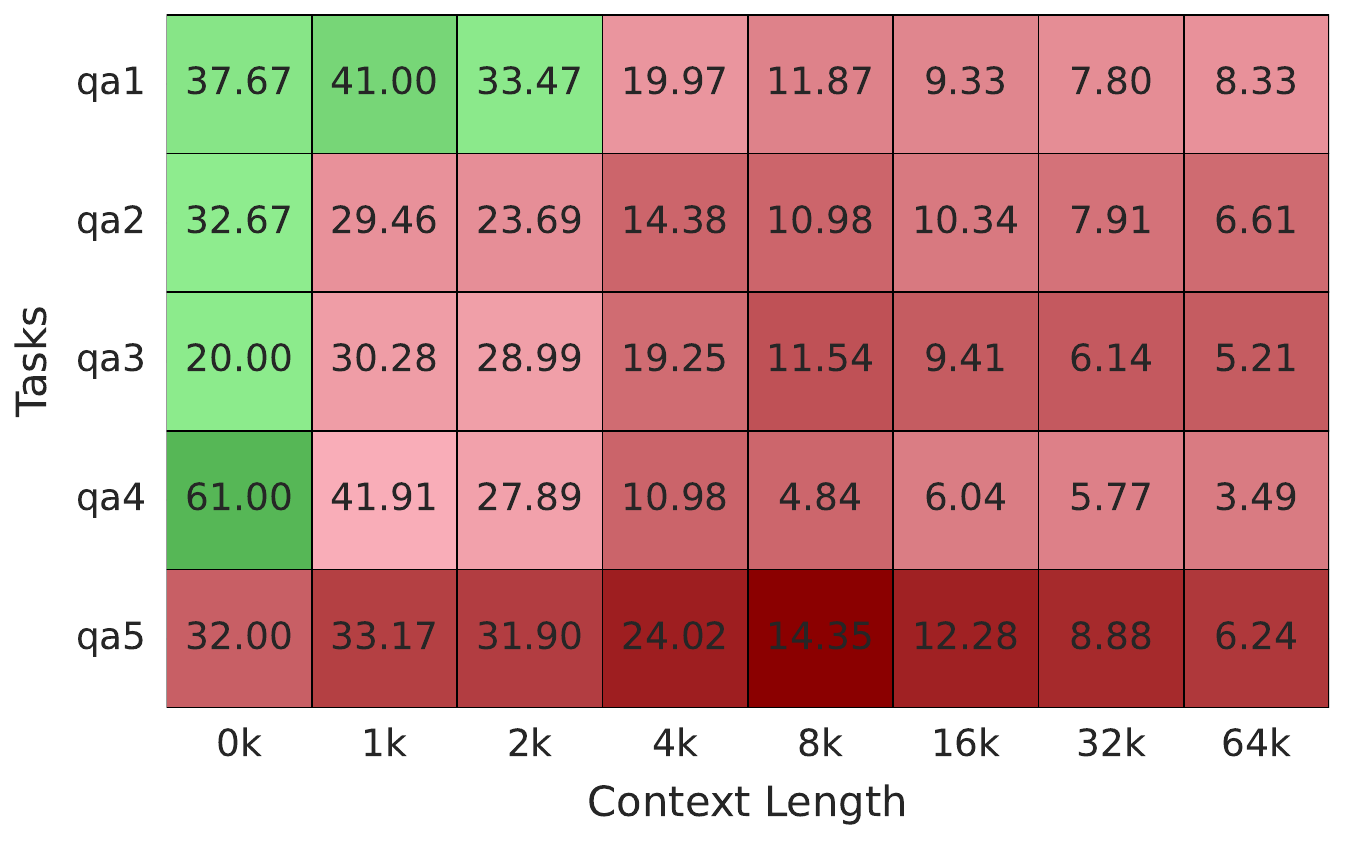}
        \caption{Mamba}
        \label{fig:subfig2_ft_mambaApp}
    \end{subfigure}
    
    \caption{Comparison of original scores of Diff-Mamba and Mamba models after fine-tuning on BABILong. Values are the percentage of correct examples, while green indicates a larger score than the other model, and red indicates the opposite.}
    \label{fig:niddle_finetuned}
\end{figure}

The models were fine-tuned on 90\% examples up to 1k tokens inside the dataset across all tasks, which is approximately equivalent to 17k examples. The other 10\% remained for test. Following~\citep{NEURIPS2024_c0d62e70}, we employ a preprocessing step that shapes each sample as follows: "\textbf{\texttt{<context>\{input\}</context> Question:\{question\} Answer:}}" and the loss was calculated on the answer label only. Since the training tiny size, three seeds have been tested and averaged through the results.
Original scores for each model are presented in Figure \ref{fig:niddle_finetuned}. The configuration for the training is in Table \ref{tab:hyperparameters_babilong}.

\begin{table}[!h]
    \centering
    \vspace{10pt}
    \small
    \begin{tabular}{@{\extracolsep{\fill}}lc}
        \toprule
        Params & Values \\
        \midrule
        Layers & 48 \\
        lr & 3e-4 \\
        Max Length & 2048 \\
        Steps & 500 \\
        Batch Size  & 6 \\
        Warmup Steps & 50 \\
        Optimizer & AdamW \\
        Weight Decay & 0.1 \\
        \bottomrule\\
    \end{tabular}
     \caption{Hyperparameters used for the BABILong fine-tuning\label{tab:hyperparameters_babilong}}
\end{table}

\subsection{Noise in Intermediate Representations}
To measure the signal-to-noise ratio, we employ Tuned-lens~\citep{belrose2023eliciting}. Specifically, we train linear projections to learn the appropriate transformation over intermediate representations across layers and we demonstrate it using the BABILong finetuned models from Section~\ref{subsec:resRetrieval}. The probe for finetuned BABILong models have been trained on the validation set of The Pile for 3 seeds with a similar configuration to the original Tuned-lens paper. Evaluation was conducted on a test set derived from the BABILong benchmark, consisting of 1,000 examples with context lengths ranging from 1k to 2k tokens. These examples were sampled from the first five tasks, with 200 examples per task. For each example, the needle probability was extracted from every layer.

\section{Efficiency Benchmark}
\label{sec:efficiency}
We evaluate the computational efficiency of Diff-Mamba-Hybrid (375M params) compared to Mamba (368M params) with respect to inference speed, memory footprint, and forward-pass latency. These measurements provide a comprehensive view of the trade-offs introduced by the architectural modifications.

\subsection{End-to-End Inference}
\begin{figure}
    \centering
    \includegraphics[width=\linewidth]{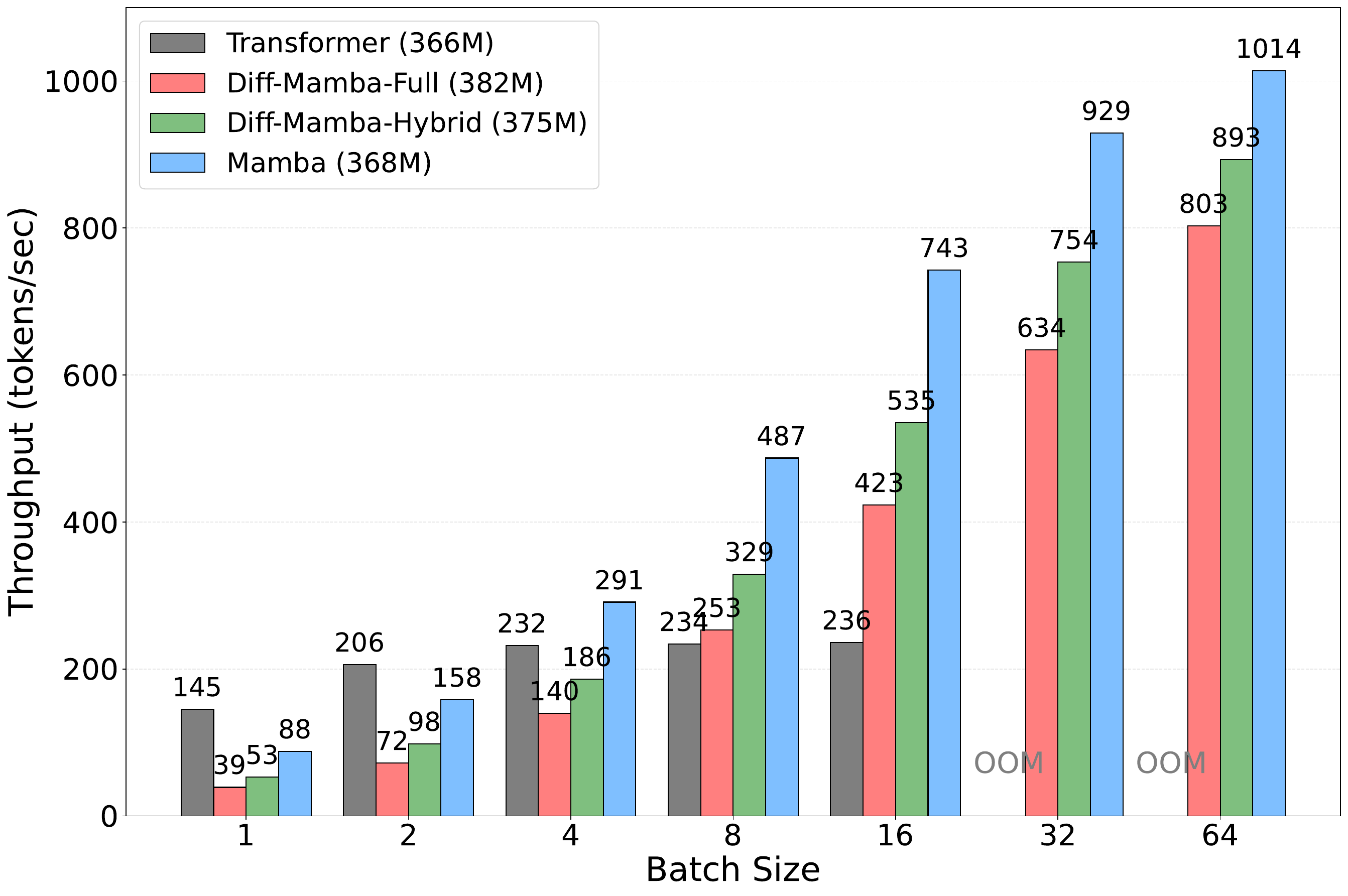}
    \vspace{-20pt}
    \caption{Inference throughput on L40s 48GB (prompt length 2048).}
    \label{fig:inference_throughput}
\end{figure}

We calculate end-to-end inference throughput across varying batch sizes, measured on an L40S GPU with 48GB of memory (Figure~\ref{fig:inference_throughput}). The benchmark was conducted using a prompt length of 2048 tokens and generating 128 new tokens with untrained models of Mamba, Diff-Mamba-Full, Diff-Mamba-Hybrid, and Transformer for a baseline. Throughput (tokens/s) is computed as \emph{batch size $\times$ 128 / inference time}. Both fully and hybrid Diff-Mamba exhibits lower throughput compared to the original Mamba at small batch sizes; however, this gap narrows considerably as the batch size increases, and becomes only a 12\% difference with a batch size of 64 for Diff-Mamba-Hybrid.

\subsection{Memory Benchmark}
GPU memory usage across different batch sizes is reported in Table~\ref{tab:memory_usage}. Each batch consists of sequences of length 2048, and memory was measured after a forward pass. Diff-Mamba incurs only a modest increase in memory consumption relative to Mamba, with a difference of less than 1GB even at batch size 32 (44.30GB vs. 43.80GB). This result suggests that the additional computational layers introduced in Diff-Mamba achieve efficiency improvements without imposing prohibitive memory demands, thereby maintaining practical feasibility in resource-constrained environments.

\begin{table}[h]
\centering
\small
\begin{tabular}{lcc}
\toprule
{Batch Size} & {Mamba (GB)} & {Diff-Mamba (GB)} \\
\midrule
1 & 3.92 & 3.95 \\
2 & 5.16 & 5.25 \\
4 & 7.72 & 7.84 \\
8 & 12.84 & 13.03 \\
16 & 23.09 & 23.43 \\
32 & 43.80 & 44.30 \\
\bottomrule
\end{tabular}
\caption{Memory usage comparison across different batch sizes.}
\label{tab:memory_usage}
\end{table}

\subsection{Forward-Pass Latency}
We further examine forward-pass latency across varying sequence lengths, summarized in Table~\ref{tab:forward_time}. Forward-pass time serves as a direct proxy for training efficiency, since it closely aligns with per-step training duration. The results demonstrate that Diff-Mamba sustains comparable per-step efficiency to Mamba, introducing only marginal latency overhead at extended sequence lengths.

\begin{table}[h]
\centering
\small
\begin{tabular}{lcc}
\toprule
{Seq Length} & {Mamba (s)} & {Diff-Mamba (s)} \\
\midrule
512 & 0.063 & 0.076 \\
1024 & 0.079 & 0.090 \\
2048 & 0.073 & 0.087 \\
4096 & 0.147 & 0.172 \\
8192 & 0.271 & 0.282 \\
\bottomrule
\end{tabular}
\caption{Forward time comparison for different sequence lengths.}
\label{tab:forward_time}
\end{table}


\end{document}